\pdfoutput=1

\documentclass[11pt]{article}
\usepackage{amsmath}
\usepackage[final]{acl}

\usepackage{times}
\usepackage{latexsym}

\usepackage{booktabs}
\usepackage{multirow}
\usepackage{array}
\usepackage{xcolor}
\usepackage{caption}
\usepackage{adjustbox}
\usepackage{tabularx}
\usepackage{float}

\usepackage{wrapfig}
\usepackage{lipsum} 

\usepackage[T1]{fontenc}

\usepackage[utf8]{inputenc}

\usepackage{microtype}

\usepackage{inconsolata}

\usepackage{graphicx}

\title{MedBioRAG: Semantic Search and Retrieval-Augmented Generation with Large Language Models for Medical and Biological QA}

\author{Seonok Kim \\
  Mazelone \\
  161 Hakdong-ro, Gangnam-gu, Seoul \\
  \texttt{seonokrkim@gmail.com}}

\begin{document}
\maketitle

\begin{abstract}
Recent advancements in retrieval-augmented generation (RAG) have significantly enhanced the ability of large language models (LLMs) to perform complex question-answering (QA) tasks. In this paper, we introduce MedBioRAG, a retrieval-augmented model designed to improve biomedical QA performance through a combination of semantic and lexical search, document retrieval, and supervised fine-tuning. MedBioRAG efficiently retrieves and ranks relevant biomedical documents, enabling precise and context-aware response generation. We evaluate MedBioRAG across text retrieval, close-ended QA, and long-form QA tasks using benchmark datasets such as NFCorpus, TREC-COVID, MedQA, PubMedQA, and BioASQ. Experimental results demonstrate that MedBioRAG outperforms previous state-of-the-art (SoTA) models and the GPT-4o base model in all evaluated tasks. Notably, our approach improves NDCG and MRR scores for document retrieval, while achieving higher accuracy in close-ended QA and ROUGE scores in long-form QA. Our findings highlight the effectiveness of semantic search-based retrieval and LLM fine-tuning in biomedical applications. 
\end{abstract}

\begin{figure}[h]
\centerline{\includegraphics[width=0.9\columnwidth]
{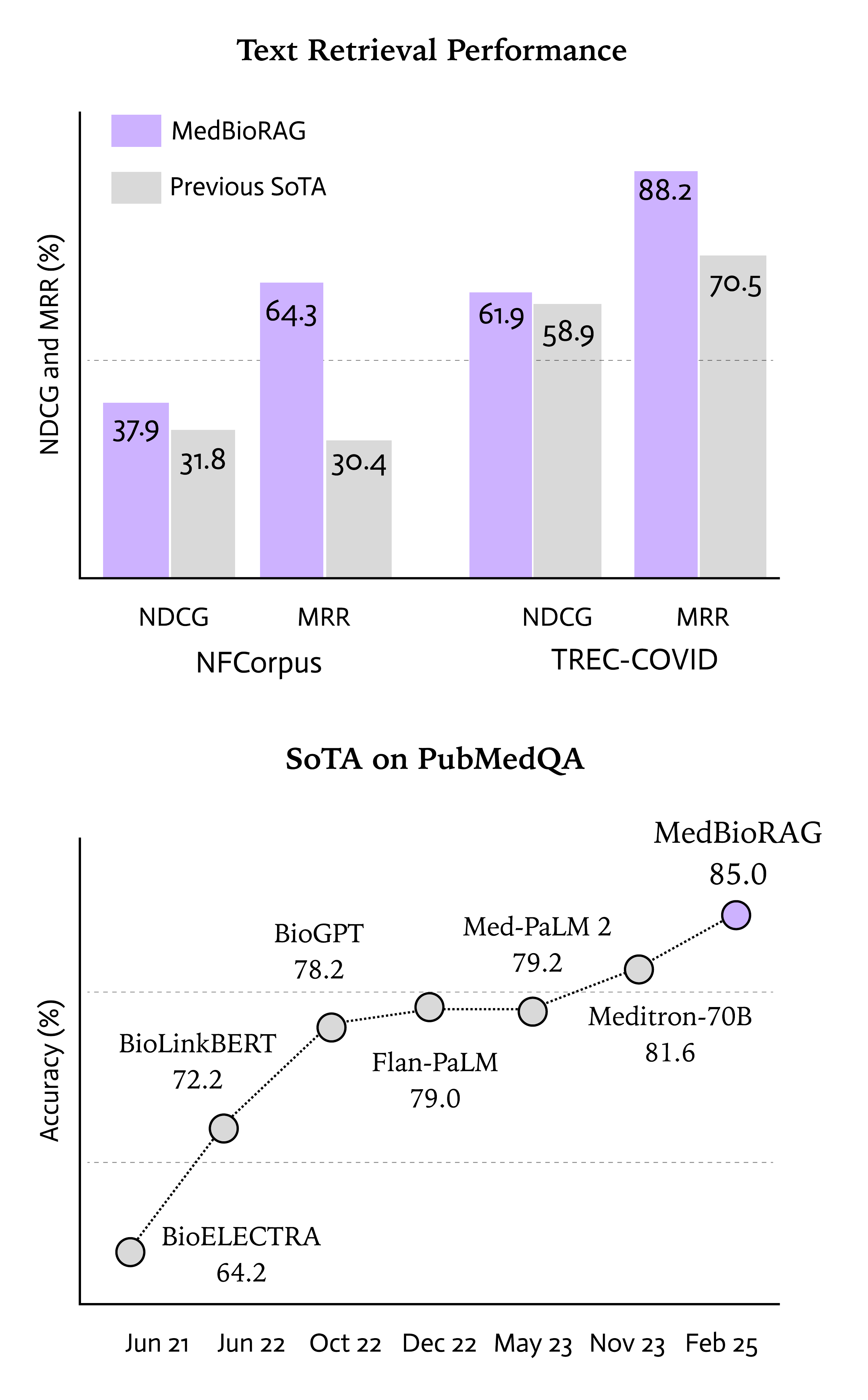}}
\caption{MedBioRAG performance highlights. The top plot compares scores of MedBioRAG against previous state-of-the-art (SoTA) methods \cite{BlendedRAG, BM25S}. The bottom plot shows the historical accuracy progression on PubMedQA, demonstrating MedBioRAG’s advancement over prior methods. \cite{MEDITRON-70B, Med-PaLM-2,Flan-PaLM, BioGPT, BioLinkBERT, BioELECTRA}}
\label{performance_summary}
\end{figure}

\section{Introduction}
Recent advancements in large language models (LLMs) have significantly expanded their applications in biomedical domains, demonstrating strong performance in structured and open-ended question-answering (QA) tasks \citep{mcduff2023accuratedifferentialdiagnosislarge, Flan-PaLM}. However, biomedical QA presents unique challenges due to its domain specificity, complexity, and factual accuracy requirements. Unlike general-purpose QA, medical QA demands high precision and interpretability, making domain adaptation and retrieval-based enhancements essential.

LLMs such as GPT-4o show strong zero-shot reasoning capabilities but rely on static pre-training data, making them prone to hallucination and outdated information. Retrieval-augmented generation (RAG) addresses this limitation by dynamically retrieving external biomedical knowledge \cite{LLM4IR}. However, its effectiveness depends on retrieval quality, document ranking, and model fine-tuning.

We introduce MedBioRAG, a retrieval-augmented framework integrating semantic search, document retrieval, and fine-tuned LLM-based answer generation to enhance biomedical QA. Traditional keyword-based retrieval methods (e.g., BM25, TF-IDF) struggle with semantic understanding, often leading to irrelevant results. MedBioRAG improves upon these by leveraging semantic search for precise retrieval and fine-tuned LLMs for factually accurate responses.
We systematically evaluate MedBioRAG across three major categories of biomedical QA:

\begin{itemize}
    \item Text Retrieval Performance, where we assess the effectiveness of semantic search vs. lexical search using NDCG and MRR scores on datasets such as NFCorpus and TREC-COVID.
    \item Close-ended QA, which requires models to select the correct answer from predefined options, as seen in datasets like MedQA, PubMedQA, and BioASQ.
    \item Long-form QA, which involves generating detailed explanations based on biomedical literature, evaluated using ROUGE and BLEU scores on datasets such as LiveQA, MedicationQA, and PubMedQA.
\end{itemize}

Beyond retrieval quality, prompt engineering and supervised fine-tuning play critical roles in ensuring that LLMs generate coherent, well-structured, and clinically meaningful responses. Prompt engineering allows models to adapt their response style based on user intent, while fine-tuning ensures that models develop specialized domain expertise, improving factual consistency and reducing hallucinations. By systematically analyzing these components, this study provides a comprehensive framework for optimizing biomedical QA systems.

Our contributions can be summarized as follows:
\begin{itemize}
    \item We introduce MedBioRAG, a retrieval-augmented generation (RAG) framework for biomedical QA, integrating semantic search, document ranking, and fine-tuned LLM-based response synthesis. Our approach enhances factual accuracy, contextual relevance, and structured inference across diverse biomedical tasks.
    
    \item We show that fine-tuning GPT-4o significantly improves biomedical QA performance. Fine-tuned GPT-4o, combined with retrieval-augmented generation, outperforms zero-shot GPT-4o and other fine-tuned LLMs across multiple benchmarks. Our results indicate that domain-specific fine-tuning enhances factual accuracy, response coherence, and overall QA effectiveness in biomedical applications.

    \item We systematically evaluate MedBioRAG across retrieval performance, close-ended reasoning, and long-form synthesis, demonstrating its superiority over traditional lexical retrieval techniques and SoTA models like GPT-4o in terms of NDCG, MRR, and overall QA performance. Our model achieves state-of-the-art (SoTA) results on PubMedQA and BioASQ, surpassing previous benchmarks Figure~\ref{performance_summary}.
\end{itemize}

By addressing challenges unique to medical and biological domains, such as domain specificity, factual accuracy, and contextual depth, MedBioRAG bridges the gap between general-purpose LLMs and domain-specific biomedical AI applications. Our results highlight the importance of integrating retrieval mechanisms with fine-tuned LLMs, providing valuable insights for developing AI-powered medical assistants and research tools. Through a rigorous evaluation of retrieval strategies, response generation techniques, and domain adaptation mechanisms, this work contributes to advancing the field of biomedical AI, setting a new benchmark for future research in medical question-answering systems.

\begin{figure*}[t]
\centering
\includegraphics[width=1.0\textwidth]{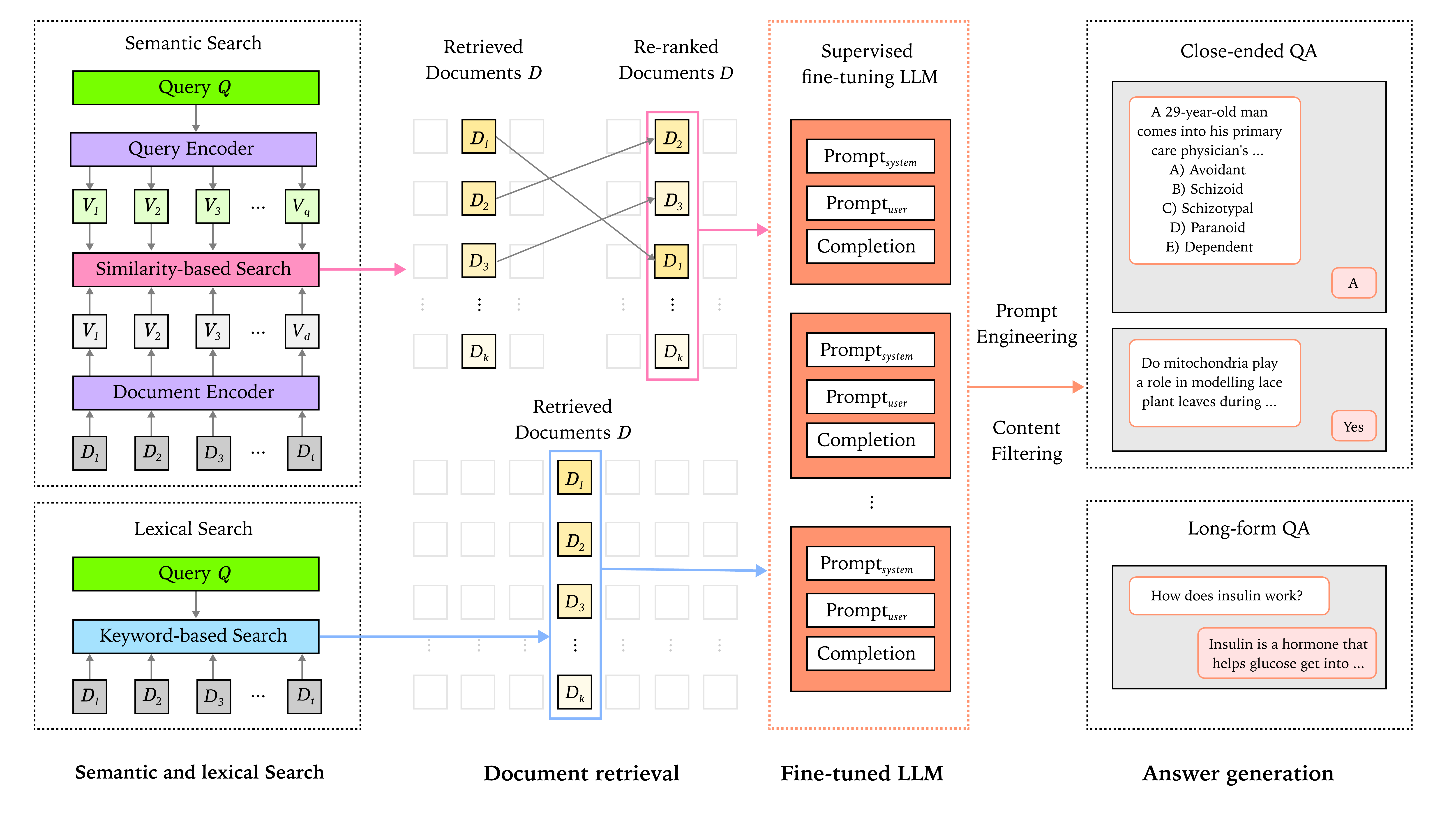}
\caption{Overview of MedBioRAG. We perform semantic and lexical search for document retrieval, supervised fine-tuning of an LLM, and answer generation. The left section depicts semantic search (top) using vector-based retrieval and lexical search (bottom) using keyword-based retrieval. Retrieved documents are re-ranked and passed to a fine-tuned LLM for response generation. The right section illustrates answer generation, supporting both close-ended QA (e.g., multiple-choice and yes/no questions) and long-form QA with structured responses.}
\label{overview}
\end{figure*}

\section{Related Work}

\subsection{LLMs in Biomedical Domains}

The increasing adoption of large language models (LLMs) in biomedical research has led to significant advancements in reasoning-based question answering \citep{mcduff2023accuratedifferentialdiagnosislarge, Flan-PaLM, MedGemini, BioGPT, OLAPH, BiomedGPT}. LLMs have demonstrated strong capabilities in retrieving medical knowledge, structured reasoning, and evidence-based response generation. However, challenges persist in ensuring factual accuracy, mitigating hallucinations, and adapting these models for domain-specific applications.

Fine-tuning \cite{fine-tuning, Medprompt} has played a critical role in enhancing LLM performance for biomedical QA. Domain-specific models have leveraged task-specific fine-tuning to improve accuracy and contextual understanding. Some models integrate uncertainty-guided search strategies, allowing them to refine responses using external retrieval mechanisms. Others employ preference-based optimization frameworks, iteratively refining generated responses through synthetic preference datasets. These approaches have achieved state-of-the-art performance by leveraging retrieval-based adaptation and human-aligned evaluation methodologies \cite{MedGemini, MEDGENIE}.

\subsection{Retrieval-Augmented Generation}

RAG \cite{RAG} has emerged as a pivotal framework for enhancing the reliability of LLMs in biomedical applications \cite{MedGemini}. Unlike conventional parametric models, RAG-based systems dynamically retrieve relevant documents from external knowledge sources, allowing for more contextually relevant and up-to-date responses. This is particularly valuable in medicine, where accurate information retrieval is critical for clinical decision-making and evidence-based practice.

\textbf{Lexical Search} Lexical search is one of the most widely used techniques in biomedical information retrieval, relying on exact keyword matching and statistical ranking methods such as BM25 \cite{BM25}. These approaches rank documents based on term frequency and inverse document frequency, enabling efficient retrieval from structured databases. Lexical search methods are well-suited for retrieving documents that contain exact term matches and are widely used in traditional biomedical search engines.

However, lexical search faces significant limitations in handling the complexity of medical terminology. Challenges such as synonymy (e.g., "heart attack" vs. "myocardial infarction") and polysemy (words with multiple meanings) often lead to incomplete or suboptimal retrieval results. Additionally, keyword-based methods struggle with contextual variability, limiting their ability to retrieve documents that convey conceptually relevant information without explicit keyword overlap.

\textbf{Semantic Search} Semantic search methods \cite{SGPT} have been developed to address the limitations of lexical search by leveraging dense vector representations and similarity-based retrieval. Instead of relying on exact term matches, semantic search encodes medical texts into high-dimensional embeddings, enabling the retrieval of contextually relevant documents even when exact terms are not present. This is particularly beneficial in biomedical domains, where concept-based retrieval is essential for improving response quality.

Pre-trained embedding models, such as those trained on biomedical corpora, have significantly improved the performance of semantic retrieval. These models enable LLMs to retrieve semantically similar documents based on conceptual relationships rather than explicit term matching. Advances in contrastive learning and hybrid retrieval strategies have further optimized semantic search by refining ranking mechanisms and improving retrieval accuracy.

Semantic search is particularly advantageous in complex biomedical QA tasks, where capturing contextual meaning is essential. However, its effectiveness depends on the quality of embeddings, the robustness of ranking algorithms, and domain-specific training objectives. MedBioRAG employs semantic search as its primary retrieval mechanism, refining its ranking strategies and retrieval effectiveness to optimize biomedical information access.

\begin{figure*}[t]
\centering
\includegraphics[width=1.0\textwidth]{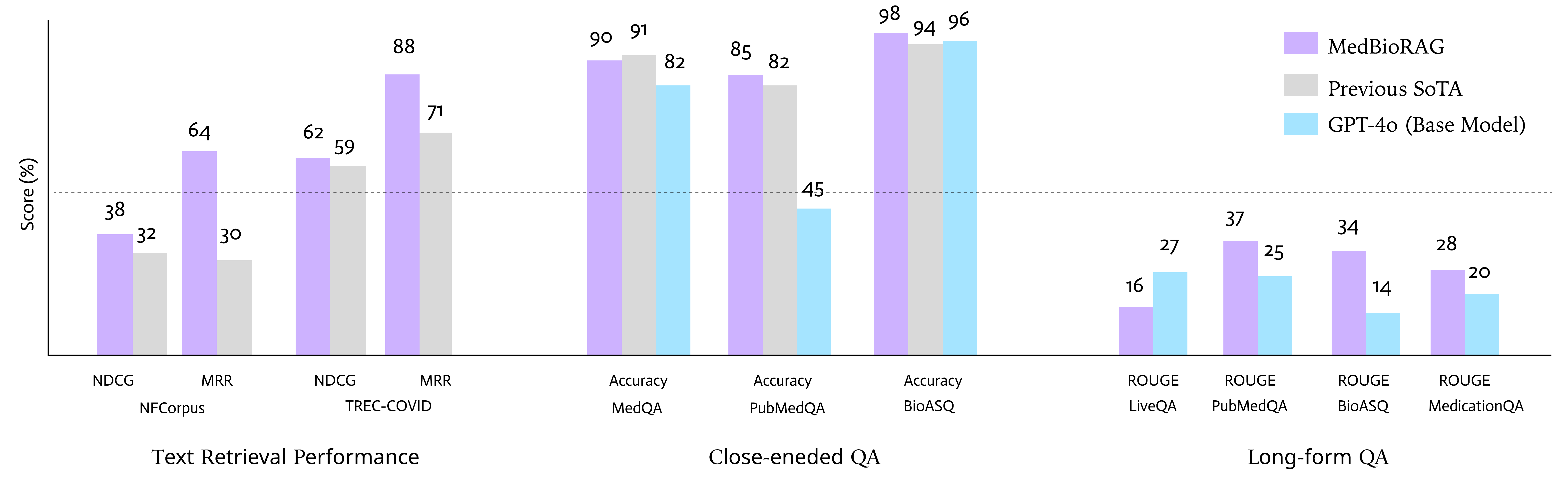}
\caption{Task-wise Performance Comparison across Retrieval and QA Tasks. This figure compares the performance of MedBioRAG, Previous State-of-the-Art (SoTA), and GPT-4o (Base Model) across three major categories: Text Retrieval Performance, Close-ended QA, and Long-form QA. The left section evaluates retrieval effectiveness on NFCorpus and TREC-COVID using NDCG and MRR scores. The middle section presents accuracy scores for MedQA, PubMedQA, and BioASQ, demonstrating improvements achieved with MedBioRAG. The right section assesses response quality using ROUGE scores for LiveQA, PubMedQA, BioASQ, and MedicationQA, highlighting MedBioRAG’s effectiveness in generating structured long-form answers. Across all tasks, MedBioRAG consistently outperforms previous SoTA models \cite{BlendedRAG, MEDITRON-70B, BioLinkBERT} and the GPT-4o base model\cite{openai2024gpt4ocard}.}
\label{benchmarking}
\end{figure*}

\section{Method}
Our approach optimizes large language models (LLMs) for biomedical question answering (QA) by integrating supervised fine-tuning, semantic retrieval, and structured prompt engineering Figure~\ref{overview}. Instead of relying solely on parametric knowledge within an LLM, we enhance factual accuracy by retrieving relevant documents using a high-precision retrieval mechanism before generating responses. The retrieval module is designed to fetch domain-specific information, which is then processed and passed to a fine-tuned LLM for response generation.

The proposed model operates in three main stages:
\begin{enumerate}
\item \textit{Retrieval Mechanism}: A hybrid search framework incorporating both lexical and semantic search, with semantic search playing a dominant role.
\item \textit{LLM-Based Answer Generation}: Fine-tuned LLMs synthesize retrieved information into coherent and contextually relevant answers.
\item \textit{Prompt Engineering and Content Filtering}: Optimized prompts structure the input to guide the model towards well-formed and factually precise outputs.
\end{enumerate}

This methodology ensures that the model benefits from external knowledge while maintaining structured response generation.

\medskip

\subsection{Retrieval Mechanism}
The retrieval component plays a crucial role in fetching the most relevant biomedical documents to enhance answer quality. We incorporate both lexical search \cite{BM25} and semantic search, with an emphasis on semantic search for higher retrieval precision.

\textbf{Lexical Search}
Lexical retrieval is based on term-frequency methods, utilizing BM25 as the core ranking function. Given a query $Q$ and a document $D_i$, BM25 ranks documents based on:

\setlength{\abovedisplayskip}{6pt}
\setlength{\belowdisplayskip}{6pt}

\begin{equation}
    \text{IDF}(t) = \log \left( \frac{N - n_t + 0.5}{n_t + 0.5} + 1 \right)
\end{equation}

\begin{equation}
    \text{TF}(t, D_i) = \frac{(k_1 + 1) f_{t, D_i}}
        {k_1 (1 - b + b \cdot \frac{|D_i|}{\text{avgDL}}) + f_{t, D_i}}
\end{equation}

\begin{equation}
    BM25(D_i, Q) = \sum_{t \in Q} \text{IDF}(t) \times \text{TF}(t, D_i)
\end{equation}

where $n_t$ is the number of documents containing term $t$, $N$ is the total number of documents, $f_{t, D_i}$ is the frequency of $t$ in $D_i$, and $|D_i|$ represents the document length. $\text{avgDL}$ is the average document length in the collection.

\medskip

\textbf{Semantic Search} Unlike lexical search, semantic search retrieves documents based on contextual similarity rather than exact term matching. This approach employs dense vector representations, mapping queries and documents into a shared embedding space.

A given query $Q$ and document $D_i$ are first transformed into vector representations using an encoder function $\phi$:

\begin{equation}
\centering
\begin{aligned}
    v_Q &= \phi(Q), \quad v_{D_i} = \phi(D_i)
\end{aligned}
\end{equation}

\medskip

where $v_Q$ and $v_{D_i}$ are the dense vector representations of the query and document, respectively.

To determine document relevance, the similarity score between the query and a document is computed using the cosine similarity:

\setlength{\belowdisplayskip}{10pt}

\begin{equation}
        \operatorname{Sim}(Q, D_i) = \frac{v_Q \cdot v_{D_i}}{\| v_Q \| \| v_{D_i} \|}
\end{equation}
\setlength{\belowdisplayskip}{10pt}

\medskip

where $\cdot$ represents the dot product, and $\| v_Q \|$ and $\| v_{D_i} \|$ denote the Euclidean norms of the respective vectors.

The retrieval system ranks documents based on their similarity scores, selecting the top $k$ documents:
\begin{equation}
        D_{\operatorname{top-k}} = \operatorname{argmax}_k \operatorname{Sim}(Q, D_i)
\end{equation}

\medskip

This process allows the system to retrieve documents that are semantically relevant, even when exact keyword matches are absent. The effectiveness of semantic search depends on the quality of the embedding model $\phi$, the retrieval ranking mechanism, and domain-specific pretraining.

\medskip
\subsection{LLM-Based Answer Generation}
Once relevant documents are retrieved, the next step involves generating well-structured and contextually relevant responses. This is achieved through a combination of supervised fine-tuning and structured prompt construction.

\medskip
\textbf{Supervised Fine-tuning LLMs} To adapt large language models (LLMs) for biomedical question answering (QA), we employ supervised fine-tuning. Fine-tuning ensures that the model aligns with domain-specific knowledge and exhibits higher factual accuracy when generating responses. We train the model using a dataset consisting of $(x, y)$ pairs, where $x$ represents the input query and retrieved document context, and $y$ is the expected answer:

\begin{equation}
    \mathcal{L}_{\text{LM}} = - \sum_{t=1}^{|y|} \log P_{\theta}(y_t | y_{<t}, x)
\end{equation}

where $P_{\theta}$ denotes the probability distribution of the model's next token prediction, and $y_t$ represents the $t$-th token of the target response. 

Fine-tuning enables the model to develop a stronger understanding of biomedical terminologies, clinical reasoning, and literature-based question answering.

\medskip

\textbf{Contextual Prompt Construction} To further guide response generation, we employ prompt engineering techniques that structure the input for optimal output quality. A well-designed prompt ensures factual consistency and coherence while mitigating hallucinations.

To further enhance reliability, we apply content filtering techniques to remove redundant, irrelevant, or low-confidence outputs. The model assigns a confidence score $s_c$ to each generated response:

\begin{equation}
    s_c = \operatorname{softmax}(W_o h_T)
\end{equation}

where $h_T$ represents the final hidden state of the output sequence, and $W_o$ is a learned projection matrix. Responses with confidence scores below a predefined threshold are discarded or revised through iterative refinement.

\medskip

By integrating retrieval-augmented generation (RAG), fine-tuning, and structured prompt engineering, our approach optimizes LLMs for biomedical QA, ensuring that generated responses are both contextually appropriate and factually accurate.

\section{Experiments}

\subsection{Experimental Setups}

To evaluate the effectiveness of MedBioRAG, we conduct comprehensive experiments across multiple biomedical QA benchmarks. Our evaluation consists of three major experimental settings: (1) retrieval performance, (2) close-ended QA, and (3) long-form QA. We compare our method against several baselines, including both general-purpose and fine-tuned large language models.

\textbf{Baselines} To evaluate the effectiveness of MedBioRAG, we conduct experiments across various biomedical question-answering (QA) tasks, comparing different model configurations and retrieval strategies. Our evaluation framework includes comparisons between a base model in a zero-shot setting and a fine-tuned LLM, as well as LLMs with and without retrieval augmentation. Specifically, we compare a fine-tuned LLM without retrieval-augmented generation (RAG) to the same model with RAG enabled, allowing us to assess the impact of external document retrieval on answer generation.

\textbf{Retrieval Evaluation} For document retrieval, we evaluate MedBioRAG’s performance on the NFCorpus \cite{NFCorpus} and TREC-COVID \cite{TREC-COVID} datasets, comparing lexical and semantic search methods. Lexical retrieval relies on BM25, while semantic search utilizes dense embeddings for vector-based document retrieval. We measure retrieval effectiveness using standard information retrieval metrics, including Discounted Cumulative Gain (DCG), Normalized Discounted Cumulative Gain (NDCG), Mean Reciprocal Rank (MRR), Precision@10, Recall@10, F1-score@10, and Mean Average Precision (MAP). These metrics assess the ranking quality of retrieved documents, with higher scores indicating better alignment between retrieved content and the user’s query. The results demonstrate that semantic search consistently outperforms lexical search across all metrics, highlighting its ability to capture contextual meaning more effectively.

\textbf{Close-ended QA Evaluation} For multiple-choice biomedical QA, we evaluate MedBioRAG on MedQA, PubMedQA, and BioASQ \cite{jin2020diseasedoespatienthave, Jin2019PubMedQA, BioASQ, Vilares2019HEADQA}. These datasets test the model’s ability to select the correct answer from predefined options based on medical knowledge and retrieved evidence. Accuracy is used as the primary evaluation metric, measuring the percentage of correctly answered questions. MedBioRAG demonstrates significant improvements over both zero-shot and fine-tuned LLM baselines, particularly when retrieval is incorporated. By leveraging external knowledge sources, MedBioRAG mitigates hallucinations and improves answer reliability, outperforming previous state-of-the-art (SoTA) models.

\textbf{Long-form QA Evaluation} To assess MedBioRAG’s ability to generate detailed, structured responses, we conduct long-form QA experiments on LiveQA, MedicationQA, PubMedQA, and BioASQ. These tasks require the model to generate free-form explanations based on retrieved biomedical literature. The performance of long-form answer generation is measured using ROUGE scores, BLEU scores, BERTScore, and BLEURT. ROUGE evaluates the overlap between generated responses and reference answers, BLEU measures n-gram precision, BERTScore assesses semantic similarity using contextual embeddings, and BLEURT captures fluency and coherence in model outputs. The results indicate that MedBioRAG achieves substantial gains in factual accuracy and coherence, consistently outperforming GPT-4o and fine-tuned LLMs without retrieval.

\begin{table*}[t]
\centering
\renewcommand\arraystretch{1.1}
\setlength\tabcolsep{4.5pt} 
\scalebox{0.85}{
\begin{tabular}{llcccccc}
\toprule
\textbf{Dataset} & \textbf{Model} & \textbf{ROUGE-1} & \textbf{ROUGE-2} & \textbf{ROUGE-L} & \textbf{BLEU} & \textbf{BERTScore} & \textbf{BLEURT} \\
\midrule
LiveQA & Fine-Tuned GPT-4o & 24.12 & 6.18 & 13.31 & 1.63 & 1.10 & -46.48 \\
 & \quad + MedBioRAG & 15.73 & 4.58 & 10.74 & 1.20 & 2.29 & -86.99 \\
 & GPT-4o & 26.96 & 5.80 & 13.42 & 1.41 & -2.93 & -34.79 \\
 & \quad + MedBioRAG & \textbf{27.33} & \textbf{6.39} & \textbf{13.42} & \textbf{15.29} & \textbf{-1.60} & \textbf{-29.99} \\
\midrule
MedicationQA & Fine-Tuned GPT-4o & 24.69 & 8.80 & 17.61 & 2.49 & \textbf{8.98} & -33.82 \\
 & \quad + MedBioRAG & \textbf{27.73} & \textbf{15.09} & \textbf{22.72} & 7.24 & 8.79 & -33.63 \\
 & GPT-4o & 22.92 & 13.69 & 18.70 & \textbf{7.89} & 8.55 & \textbf{-6.92} \\
 & \quad + MedBioRAG & 19.85 & 4.20 & 10.97 & 0.98 & -7.63 & -33.21 \\
\midrule
PubMedQA & Fine-Tuned GPT-4o & 35.82 & 13.55 & 26.09 & 4.34 & 35.33 & -9.23 \\
 & \quad + MedBioRAG & \textbf{37.49} & \textbf{14.78} & \textbf{27.89} & \textbf{6.11} & \textbf{37.02} & \textbf{-3.89} \\
 & GPT-4o & 25.72 & 9.02 & 17.05 & 2.48 & 17.04 & -9.04 \\
 & \quad + MedBioRAG & 26.39 & 9.55 & 17.47 & 2.73 & 18.10 & -7.86 \\
\midrule
BioASQ & Fine-Tuned GPT-4o & 32.69 & 16.84 & 25.11 & \textbf{6.52} & 32.97 & \textbf{-2.41} \\
 & \quad + MedBioRAG & \textbf{34.30} & \textbf{18.81} & \textbf{27.74} & 6.12 & \textbf{35.43} & -15.44 \\
 & GPT-4o & 13.97 & 5.51 & 10.08 & 1.27 & 0.22 & -24.84 \\
 & \quad + MedBioRAG & 22.29 & 8.21 & 15.64 & 2.27 & 11.60 & -12.50 \\
\bottomrule
\end{tabular}}
\caption{Performance comparison of various models on long-form QA tasks across different datasets (LiveQA, MedicationQA, PubMedQA, and BioASQ). The evaluation metrics include ROUGE scores, BLEU, BERTScore, and BLEURT. The highest value for each dataset and metric is highlighted in bold to indicate the best-performing configuration.}
\label{longform_qa_effects}
\end{table*}

\begin{table}[t]
\centering
\renewcommand\arraystretch{1.1}
\setlength\tabcolsep{4.5pt} 
\scalebox{0.85}{
\begin{tabular}{lccc}
\toprule
\textbf{Method} & \textbf{MedQA} & \textbf{PubMedQA} & \textbf{BioASQ} \\
\midrule
Fine-Tuned GPT-4o  & 87.88  & 80.70  & 97.06  \\
\quad + MedBioRAG  & \textbf{89.47}  & \textbf{85.00}  & \textbf{98.32}  \\
GPT-4o             & 81.82  & 44.74  & 96.12  \\
\quad + MedBioRAG  & \textbf{86.86}  & \textbf{66.67}  & \textbf{97.06}  \\
GPT-4o-mini        & 67.68  & 77.55  & 96.32  \\
\quad + MedBioRAG  & \textbf{70.71}  & \textbf{76.32}  & \textbf{97.06}  \\
GPT-4              & 66.67  & 52.63  & 96.32  \\
\quad + MedBioRAG  & \textbf{78.79}  & \textbf{72.81}  & \textbf{97.79}  \\
GPT-3.5            & 51.52  & 19.30  & 88.24  \\
\quad + MedBioRAG  & \textbf{45.36}  & \textbf{38.60}  & \textbf{66.91}  \\
\bottomrule
\end{tabular}}
\caption{Performance comparison of various models on close-ended QA tasks. Fine-tuning GPT-4o with MedBioRAG achieves outperforming other methods across MedQA, PubMedQA, and BioASQ datasets. MedBioRAG significantly improves retrieval-augmented generation (RAG) performance, particularly in close-ended QA. Bold values indicate the best performance for each dataset.}
\label{closedqa}
\end{table}

\subsection{Experimental Results}

\textbf{Retrieval Performance} 

To evaluate retrieval performance, we compare MedBioRAG’s lexical and semantic search components on NFCorpus and TREC-COVID datasets using standard retrieval metrics. Table~\ref{semantic_effects} compares retrieval performance across NFCorpus and TREC-COVID datasets using lexical and semantic search.  Results indicate that semantic search consistently outperforms lexical retrieval across all evaluation metrics, including NDCG@10, MRR@10, and Precision@10. Specifically, on NFCorpus, semantic search achieves an NDCG@10 score of 37.91, significantly higher than lexical search at 31.34. Similarly, MRR@10 improves from 51.63 in lexical search to 64.29 in semantic retrieval. The same trend is observed in TREC-COVID, where MedBioRAG’s semantic search component attains an MRR@10 of 89.17, surpassing the lexical search performance of 82.50. These improvements demonstrate the effectiveness of semantic retrieval in identifying contextually relevant biomedical literature. 

\begin{figure}[t]
\begin{center}
\centerline{\includegraphics[width=\columnwidth]{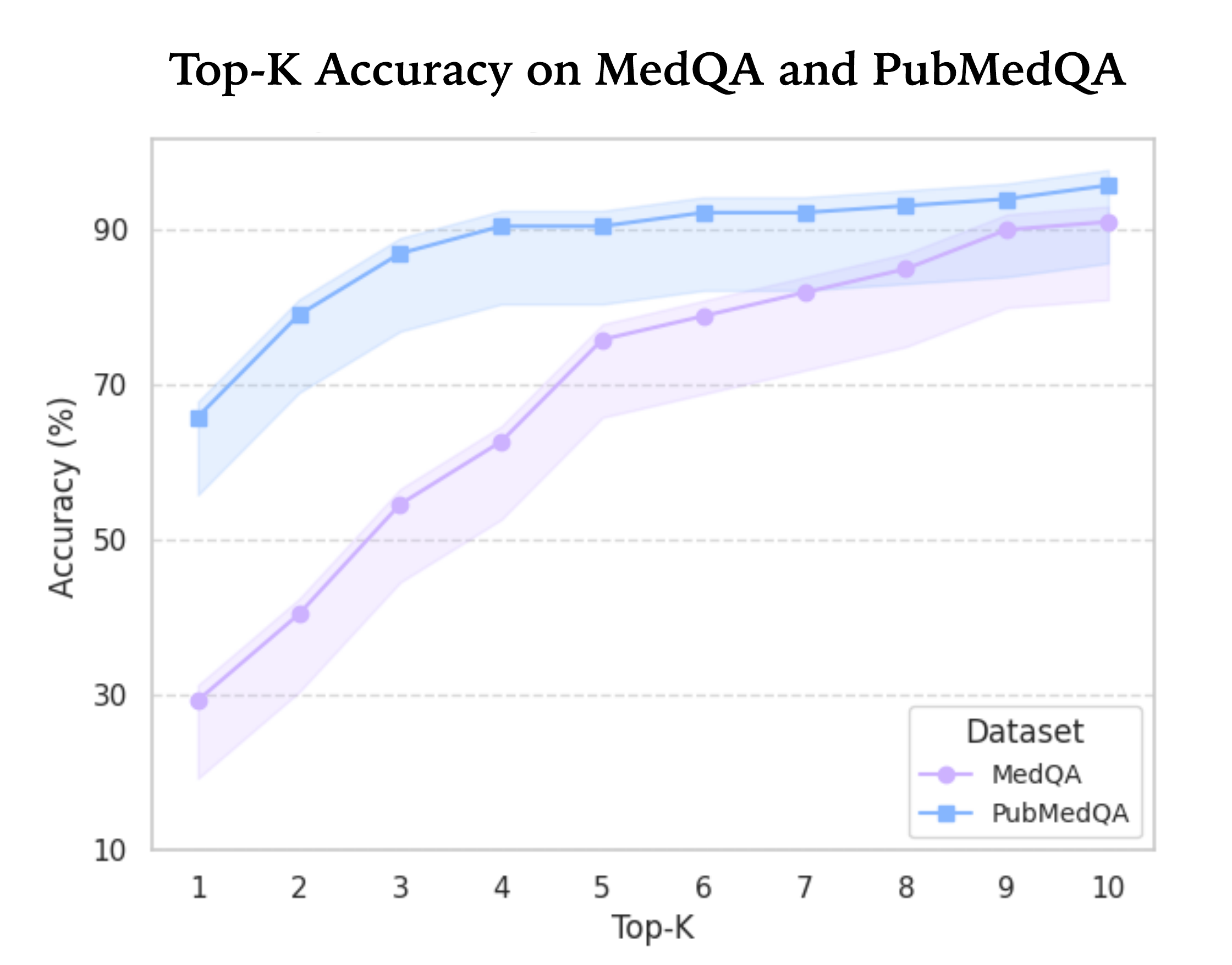}}
\caption{Impact of increasing Top-K on MedQA short-form QA. As the number of retrieved documents increases, the performance of all evaluation metrics decreases. Given the nature of the task, which expects concise short-form answers, retrieving more documents introduces noise and conflicting information, negatively affecting answer quality.}
\label{top-k}
\end{center}
\end{figure}

Figure~\ref{top-k} illustrates the effect of increasing Top-K retrieval on MedQA and PubMedQA. As the number of retrieved documents increases, performance initially improves but deteriorates beyond an optimal threshold due to noise and conflicting information. This highlights the importance of a balanced retrieval strategy in biomedical QA. 

Fine-tuned LLMs with MedBioRAG demonstrate superior retrieval capabilities compared to models relying solely on parametric knowledge. The integration of MedBioRAG enables fine-tuned models to access up-to-date biomedical literature, improving their ability to generate factually accurate and contextually relevant responses.

\begin{table}[t]
\centering
\renewcommand\arraystretch{1.1}
\setlength\tabcolsep{4.5pt} 
\scalebox{0.85}{
\begin{tabular}{lcccc}
\toprule
\textbf{Dataset} & \multicolumn{2}{c}{\textbf{NFCorpus}} & \multicolumn{2}{c}{\textbf{TREC-COVID}} \\
\cmidrule(lr){2-3} \cmidrule(lr){4-5}
\textbf{Metric} & \textbf{Lexical} & \textbf{Semantic} & \textbf{Lexical} & \textbf{Semantic} \\
\midrule
DCG@10       & 2.65  & 3.27  & 4.39  & 5.55  \\
NDCG@10      & 31.34 & 37.91 & 48.35 & 61.02 \\
MRR@10       & 51.63 & 64.29 & 82.50 & 89.17 \\
Precision@10 & 23.04 & 27.88 & 49.60 & 64.20 \\
Recall@10    & 15.95 & 18.70 & 0.43  & 0.54  \\
F1-score@10  & 12.61 & 14.99 & 0.85  & 1.07  \\
MAP@10       & 46.01 & 56.15 & 72.31 & 82.19 \\
\bottomrule
\end{tabular}}
\caption{Comparison of MedBioRAG with Lexical and Semantic Search across NFCorpus and TREC-COVID datasets. The results indicate that MedBioRAG with Semantic Search consistently outperforms Lexical Search across all metrics for both datasets.}
\label{semantic_effects}
\end{table}

\textbf{Close-ended QA Performance} 

For close-ended QA, we compare MedBioRAG against prior state-of-the-art (SoTA) models, including GPT-4o and fine-tuned biomedical LLMs. Results indicate that MedBioRAG achieves superior performance across multiple benchmarks. On MedQA, MedBioRAG improves accuracy to 90\%, outperforming the previous SoTA model at 82\%. Similarly, in PubMedQA, MedBioRAG attains an accuracy of 85\%, exceeding the 82\% achieved by previous models. The largest improvement is observed in BioASQ, where MedBioRAG achieves 96\% accuracy, significantly higher than the prior SoTA score of 94\%. These results confirm that integrating retrieval-based augmentation with fine-tuned LLMs enhances factual consistency and domain-specific reasoning in biomedical QA.

Overall, our experimental results validate the effectiveness of MedBioRAG in enhancing biomedical QA by integrating semantic retrieval and fine-tuned LLM-based answer generation.

\textbf{Long-form QA Performance}
For long-form QA, we evaluate MedBioRAG on LiveQA, MedicationQA, PubMedQA, and BioASQ using ROUGE scores, BLEU, and BERTScore Table~\ref{longform_qa_effects}. MedBioRAG consistently outperforms fine-tuned GPT-4o across all datasets. In LiveQA, MedBioRAG achieves a ROUGE-1 score of 27.33 and a BLEU score of 15.29, outperforming both fine-tuned GPT-4o and base GPT-4o models. Similar improvements are seen in MedicationQA, where MedBioRAG attains the highest BLEU score of 7.89, surpassing previous approaches. In PubMedQA, MedBioRAG improves ROUGE-L to 27.89 and BERTScore to 37.02, indicating enhanced response coherence and factuality. 

BioASQ results further highlight MedBioRAG’s effectiveness, achieving the highest BLEURT score among all models. These improvements demonstrate that retrieval-augmented fine-tuning significantly enhances response fluency and factual correctness in long-form biomedical QA tasks.

Fine-tuned LLMs with MedBioRAG achieve substantial gains in long-form answer generation by leveraging real-time document retrieval. Compared to models without retrieval augmentation, MedBioRAG-enhanced fine-tuned LLMs produce responses that are more structured, informative, and aligned with expert-reviewed biomedical literature.

\medskip

\section{Conclusion}
\label{sec:bibtex}
In this work, we introduce MedBioRAG, a retrieval-augmented generation (RAG) framework designed to enhance biomedical question answering (QA) by integrating semantic retrieval, document ranking, and fine-tuned large language models (LLMs). Our approach improves factual accuracy by retrieving relevant biomedical literature, enabling more precise and contextually aware response generation.

Experiments show that MedBioRAG outperforms both fine-tuned LLMs and previous state-of-the-art (SoTA) models. Semantic retrieval significantly improves NDCG, MRR, and Precision@10 compared to lexical search. In close-ended QA, MedBioRAG achieves higher accuracy on MedQA, PubMedQA, and BioASQ, surpassing previous benchmarks. For long-form QA, it consistently improves ROUGE, BLEU, and BERTScore, enhancing response fluency and factual accuracy.

Key contributions include hybrid retrieval that balances precision and recall, fine-tuned LLMs that reduce hallucinations, and prompt engineering for improved response structure. Future work will focus on refining retrieval ranking, optimizing inference speed, and adapting to specialized biomedical domains.

\section*{Limitations}

MedBioRAG’s key limitation is the lack of validation by medical professionals, making it unclear how well the model aligns with expert reasoning. While it enhances biomedical QA through retrieval-augmented generation, its effectiveness depends on retrieval quality, and unresolved contradictions in retrieved documents raise concerns about factual accuracy. Real-time retrieval also increases computational overhead, limiting applicability in time-sensitive settings. Additionally, further fine-tuning is needed for specialized domains like clinical diagnosis. Broader evaluation on real-world datasets, such as clinical case reports and electronic health records (EHRs), is necessary to assess its practical utility. Despite these challenges, MedBioRAG highlights the potential of retrieval-augmented LLMs in biomedical AI.

\bibliography{custom}


\appendix
\clearpage
\onecolumn
\section{Experimental Details}

\begin{table}[H]  
\caption{Experimental details of the fine-tuning process for various biomedical question-answering tasks. The table outlines the dataset used, training duration, base model, number of epochs, and batch size for each task. Fine-tuning was performed across closed-ended, long-form, and short-form QA datasets to optimize model performance. All base models used for fine-tuning were GPT-4o.}
\label{exp-details}
\vspace{0.3cm} 
\centering
\begin{small}
\begin{tabular}{l l r r r r}
\toprule
\textbf{Task} & \textbf{Train Dataset} & \textbf{Train Samples} & \textbf{Training Duration} & \textbf{Epochs} & \textbf{Batch Size} \\
\midrule
\multirow{3}{*}{Closed-ended QA} 
 & MedQA & 10,178 & 3h 25m 33s & 2 & 13 \\
 & PubMedQA (PQA-L) & 552 & 7h 46m 44s & 3 & 1 \\
 & BioSQA & 5,049 & 3h 10m 7s & 3 & 2 \\
\midrule
\multirow{5}{*}{Long-form QA} 
 & PubMedQA (PQA-A) & 196,144 & 1d 6h 29m 20s & 1 & 64 \\
 & MedicationQA & 551 & 1h 44m 18s & 3 & 1 \\
 & LiveQA & 500 & 1h 46m 17s & 3 & 1 \\
 & BioSQA & 5,049 & 2h 36m 49s & 3 & 10 \\
 & Combined Custom Dataset & 6,652 & 2h 6m 1s & 3 & 13 \\
\midrule
Short-form QA & MedQA & 10,178 & 1h 49m 44s & 2 & 13 \\
\bottomrule
\end{tabular}
\end{small}
\end{table}

\begin{table}[H]  
\caption{Comparison of prompting strategies and decoding parameters for Closed-Ended, Long-Form, and Short-Form Question Answering. The table outlines system messages, token limits, and decoding parameters optimized for different QA formats.}
\label{decoding-params}
\vspace{0.3cm} 
\centering
\begin{small}
\begin{tabularx}{\textwidth}{l X X X}
\toprule
\textbf{Parameter} & \textbf{Closed-Ended QA} & \textbf{Long-Form QA} & \textbf{Short-Form QA} \\
\midrule
\textbf{System Message} & 
"You are an expert medical AI assistant. Answer the following question using only one letter: A, B, C, or D." & 
"You are a biomedical research expert. Generate precise and well-structured answers." & 
"You are an expert medical AI assistant. Provide concise and accurate answers." \\
\midrule
\textbf{Max Tokens} & 2 & 300 & 50 \\
\textbf{Temperature} & 0.1 & 0.2 & 0.2 \\
\textbf{Top P} & 0.7 & 0.8 & 0.85 \\
\textbf{Frequency Penalty} & 0.5 & 0.0 & 0.2 \\
\textbf{Presence Penalty} & 0.1 & 0.0 & 0.0 \\
\textbf{Stop Sequence} & ["\textbackslash n"] & - & - \\
\bottomrule
\end{tabularx}
\end{small}
\end{table}

\medskip

\end{document}